В.В. Кромер

# ОБ ОДНОЙ ТРЕХПАРАМЕТРИЧЕСКОЙ МОДЕЛИ ТЕСТИРОВАНИЯ

Как известно, тесты, состоящие из заданий, не отобранных по степени трудности в соответствии со стандартом распределения, определяют измеряемое качество (свойство) на ранговой шкале. Переход к более сильной интервальной шкале требует параметризации заданий по степени их трудности и отбора в соответствии с заданным стандартом распределения. В классической теории тестов под степенью трудности (в современной версии термина) понимается доля испытуемых $q_j$, не справившихся с заданием $j$. Поскольку эта доля зависит от состава выборки испытуемых, задания необходимо должны параметризироваться на репрезентативной выборке испытуемых. В условиях локального тестирования в силу малочисленности генеральной совокупности (например, студентов одного потока данного года обучения) репрезентативной выборкой является вся генеральная совокупность.

Вытекающие из классической теории тестов ограничения снимаются в современных моделях тестирования, например логистических моделях Раша, Бирнбаума и моделях совместной количественной оценки трудностей заданий и уровней подготовленности испытуемых.

В модели Раша и ее обобщениях вводится понятие вероятности $P$ правильного выполнения $i$-м испытуемым $j$-го тестового задания, зависящей от параметров испытуемого и задания. Зависимость $P$ от уровня измеряемого свойства испытуемого и уровня трудности задания называется функцией успеха. В однопараметрической логистической модели Раша данная вероятность определяется лишь разностью уровней развития свойства испытуемого $\theta_i$ и уровня трудности задания $\beta_j$: $P = \dfrac{1}{1+e^{-(\theta_i-\beta_j)}}$, что и позволяет считать модель Раша однопараметрической (единственным параметром является выражение $(\theta_i - \beta_j)$).

В истории развития практически любой модели закономерно последующее усложнение ее путем увеличения количества параметров с целью улучшения степени приближения модели к действительности. При этом предыдущие параметры сохраняются, а изменение их значений в новой модели объясняется тем, что параметры прежней модели «вбирали» в себя функции отсутствующих (введенных лишь в новой модели) параметров и вследствие этого оказывались смещенными. При этом зачастую совершается методологическая ошибка следующего рода.

Все мыслимое множество параметров некой идеальной обобщенной модели (включающее в себя и еще не включенные в актуальную модель параметры) можно разделить на группы параметров в зависимости от степени их влияния на выход модели (и, соответственно, на степень точности отображения моделью действительности). При поэтапном усложнении модели путем увеличения количества включаемых в нее параметров улучшение на каждом этапе наблюдается лишь при соблюдении порядка включения новых параметров, а именно – порядок включения новых параметров должен соответствовать степени влияния новых параметров на результат. Отсюда вытекает, что включение в модель лишь одного параметра из группы примерно равновесных параметров методологически неверно, а декларируемое улучшение точности модели иллюзорно. В настоящем докладе общий принцип иллюстрируется на примере перехода от однопараметрической модели Раша к двухпараметрической модели Бирнбаума. Модель Раша, безусловно, базируется на самом релевантном (наиболее весомом) параметре – разности сил испытуемого и задания. Поскольку еще в классической теории тестов было выяснено, что задания обладают различной дифференцирующей способностью, в двухпараметрическую модель Бирнбаума также оказалась включенной дифференцирующая способность задания. В соответствии с логистической моделью Бирнбаума вероятность успеха испытуемого составляет $P = \dfrac{1}{1 + e^{-d_j\,(\theta_i - \beta_j)}}$, где $d_j$ – дифференцирующая способность задания. Однако в классической тестовой теории смысл введения понятия о дифференцирующей способности задания заключался в отбраковке заданий с низкой дифференцирующей способностью, т.е. заданий, плохо различающих сильных и слабых испытуемых.

Один из методов оценки дифференцирующей способности задания – определение коэффициента корреляции вектора задания с вектором суммы тестовых баллов по испытуемым. Отрицательное, нулевое или низкое значение этого коэффициента является индикатором необходимости удаления задания. Но в рамках классической же теории тестов можно убедиться в том, что степень корреляции вектора испытуемого с вектором суммы тестовых баллов по заданиям также варьирует в широких пределах, заходя в область отрицательных значений, что по аналогии требует введения параметра, условно могущего быть названным «дифференцирующей способностью испытуемого». Некоторым аналогом подобного параметра является известный т.н. «индекс настороженности», характеризующий правильность профиля испытуемого [3, с. 93]. В оборот «дифференцирующая способность испытуемого» введена не была по простой причине: исключение заданий с низкой дифференцирующей способностью из набора заданий в тестовой форме вполне законно и осуществляется в ходе «чистки теста» [1, с. 102–104]. О массовом исключении из выборки несостоятельных испытуемых вопрос никогда не ставился. В [1, с. 106] дается рекомендация исключать из матрицы тестовых результатов данные по наиболее несостоятельным испытуемым, при этом количество исключаемых испытуе-мых не должно превышать 5%, что конечно же не сравнимо по масштабу с количеством удаляемых при чистке теста заданий (до 50% и более от общего их количества). Удаление из классического теста части заданий не меняет основного параметра оставшихся заданий – степени их трудности. В модели Бирнбаума предполагается, что учет дифференцирующей способности заданий уточняет значения сил испытуемых и заданий, однако если окажется, что неучитываемый моделью фактор различной дифференцирующей способности испытуемых по степени влияния на результат (или вернее, по разбросу влияния на результат) сравним с аналогичным параметром по заданиям, подобное уточнение не имеет смысла, поскольку проходится лишь половина пути в нужном направлении.

Разрешение поставленного вопроса требует разработки трехпараметрической модели хотя бы в первом приближении. При разработке подобной модели мы исходили из следующих обстоятельств. Логистическая функция успеха не является единственно возможной. Одной из возможных является также нормальная огива, более прозрачно интерпретирующая параметры модели и по выходным результатам практически неотличимая от логистической функции. Нормальным аналогом логистической модели Бирнбаума является модель с функцией успеха $P = \Phi\left(d_j(\theta_i - \beta_j)\right)$, где $\theta_i$ и $\beta_j$ – соответственно сила испытуемого и сила задания, а $d_j$ – дифференцирующая способность задания. $\Phi$ – интегральная функция нормального распределения. Форма записи $P = \Phi\left(\dfrac{\theta_i - \beta_j}{\dfrac{1}{d_j}}\right)$ позволяет интерпретировать $\dfrac{1}{d_j}$ как некое значение $\sigma_j$, имеющее размерность силы (испытуемого или задания), а взаимоотношение между испытуемым и заданием рассматривать в рамках модели Гутмана (интерпретируемой здесь как частный случай модели Бирнбаума для заданий с бесконечной дифференцирующей способностью) при случайном изменении силы задания по нормальному закону распределения, характеризующемуся стандартным отклонением $\sigma_j$. Но ведь сила испытуемого также может испытывать случайные изменения, и охарактеризовать их можно значением стандартного отклонения силы $\sigma_i$. Изложенные обстоятельства были известны уже в классической теории тестирования, где измеренное значение тестового балла $X$ полагалось равным сумме истинного тестового балла $T$ и двух членов $e_1$ и $e_2$, характеризующих соответственно ошибку за счет изменения состояния испытуемого и ошибку, связанную с недостаточным качеством теста: $X = T + e_1 + e_2 + \ldots$ . Многоточием обозначены другие возможные ошибки (подсказка, забывание, догадка и пр.). [1, с. 60]. Комплексный учет случайных изменений си-лы как испытуемых, так и задания, позволяет сконструировать искомую трехпараметрическую модель. При независимости случайных изменений силы испытуемого и силы задания (это наиболее простой случай, с которого и необходимо начинать изучать возможностей предлагаемой модели) дисперсии (т.е. квадраты стандартных отклонений) случайных изменений сил складываются, и суммарное стандартное отклонение составит $\sigma_s = \sqrt{\left(\frac{1}{d_i}\right)^2 + \left(\frac{1}{d_j}\right)^2}$, откуда $d_s = \frac{1}{\sigma_s} = \frac{d_i d_j}{\sqrt{d_i^2 + d_j^2}}$, где за $d_s$ обозначена результирующая дифференцирующая способность задания и испытуемого. В предлагаемой трехпараметрической модели функция успеха имеет вид $P = \Phi\left(\frac{d_i d_j}{\sqrt{d_i^2 + d_j^2}}(\theta_i - \beta_j)\right)$. Возможна также и логистическая трехпараметрическая модель с функцией успеха $P = \frac{1}{1 + e^{-\frac{d_i d_j}{\sqrt{d_i^2 + d_j^2}}(\theta_i - \beta_j)}}$. Параметры трехпараметрической функции находятся применением метода максимального правдоподобия путем проведения ряда итераций, при этом в качестве начального приближения целесообразно использовать значения $\theta_i$ и $\beta_j$, полученные исходя из решения системы уравнений для однопараметрической модели, а в качестве $d_i$ и $d_j$ – значения $\sqrt{2} \approx 1{,}41$, что дает $d_s = \frac{d_i d_j}{\sqrt{d_i^2 + d_j^2}} = 1$, т.е. значение, при котором модель Бирнбаума вырождается в модель Раша.

    Модель исследовалась на результатах тестирования 46 испытуемых 44 заданиями итогового теста по немецкому языку как второму иностранному для студентов 4 курса факультета иностранных языков Новосибирского госпедуниверситета. В таблице 1 приведены рассчитанные значения силы испытуемых согласно модели Раша, модели Бирнбаума и предложенной трехпараметрической модели. Расчеты параметров модели Бирнбаума велись в двух вариантах: вариант 1 – традиционная модель Бирнбаума, где под $d$ понимается дифференцирующая способность задания; вариант 2 – под $d$ понимается дифференцирующая способность испытуемого. Для представления в таблице испытуемые ранжированы по убыванию их силы в рамках предложенной модели. Поскольку в качестве функции успеха принята нормальная огива, единицей измерения силы является не логит, как в логистической модели, а пробит – более крупная (в 1,7 раза) единица. Значения силы по всем четырем моделям получены на разных шкалах, в общем случае взаимно не центрированных и разномасштабных,

но связанных линейным преобразованием, что позволяет вычислять коэффициенты корреляции Пирсона между рядами величин.

Вычисленные значения сил испытуемых и заданий приведены к единой шкале путем линейного преобразования (с различными параметрами для каждой шкалы). Параметры линейного преобразования находятся исходя из условия равенства среднего значения сил нулю и единичного стандартного отклонения по выборке.

*Таблица 1*

**Значения сил испытуемых в рамках рассматриваемых моделей**

| № испытуемого | Модель Раша | Модель Бирнбаума | | Предлагаемая модель |
|---|---|---|---|---|
| | | Вариант 1 | Вариант 2 | |
| **1** | **2** | **3** | **4** | **5** |
| 1 | 2,10 | 2,25 | 2,32 | 2,09 |
| 2 | 2,14 | 2,14 | 2,41 | 1,97 |
| 3 | 1,60 | 1,67 | 1,77 | 1,82 |
| 4 | 1,21 | 0,90 | 1,70 | 1,63 |
| 5 | 1,90 | 1,79 | 1,13 | 1,31 |
| 6 | 1,20 | 1,27 | 1,03 | 1,27 |
| 7 | 0,89 | 0,99 | 0,89 | 1,04 |
| 8 | 0,65 | 0,90 | 0,72 | 0,95 |
| 9 | 0,87 | 0,85 | 0,82 | 0,82 |
| 10 | 0,91 | 0,76 | 0,47 | 0,78 |
| 11 | 0,76 | 0,73 | 0,66 | 0,78 |
| 12 | 0,77 | 0,69 | 0,62 | 0,74 |
| 13 | 0,34 | 0,42 | 0,24 | 0,37 |
| 14 | 0,21 | 0,37 | 0,21 | 0,37 |
| 15 | 0,19 | 0,18 | 0,07 | 0,34 |
| 16 | 0,50 | 0,23 | 0,44 | 0,24 |
| 17 | 0,09 | 0,12 | 0,06 | 0,17 |
| 18 | 0,19 | 0,14 | 0,12 | 0,12 |
| 19 | 0,06 | 0,26 | 0,03 | 0,10 |
| 20 | -0,07 | 0,18 | -0,07 | 0,09 |
| 21 | 0,08 | 0,07 | 0,11 | 0,07 |
| 22 | 0,19 | -0,01 | 0,14 | 0,06 |
| 23 | 0,19 | -0,01 | 0,05 | 0,05 |
| 24 | -0,07 | -0,06 | -0,08 | -0,01 |
| 25 | -0,04 | -0,15 | -0,05 | -0,05 |
| 26 | 0,07 | -0,06 | -0,01 | -0,06 |
| 27 | -0,08 | -0,01 | -0,08 | -0,17 |
| 28 | -0,07 | -0,24 | -0,08 | -0,17 |
| 29 | -0,19 | -0,14 | -0,20 | -0,27 |

| 1 | 2 | 3 | 4 | 5 |
|---|---|---|---|---|
| 30 | -0,62 | -0,44 | -0,45 | -0,29 |
| 31 | -0,20 | -0,56 | -0,16 | -0,29 |
| 32 | -0,33 | -0,31 | -0,29 | -0,32 |
| 33 | -0,08 | -0,12 | -0,09 | -0,33 |
| 34 | -0,47 | -0,63 | -0,47 | -0,56 |
| 35 | -0,63 | -0,79 | -0,53 | -0,64 |
| 36 | -0,77 | -0,77 | -0,54 | -0,75 |
| 37 | -0,79 | -0,78 | -0,74 | -0,79 |
| 38 | -1,07 | -0,74 | -0,72 | -0,82 |
| 39 | -1,03 | -0,81 | -0,84 | -0,97 |
| 40 | -1,06 | -1,24 | -0,85 | -1,04 |
| 41 | -1,35 | -1,11 | -1,03 | -1,10 |
| 42 | -0,76 | -0,35 | -1,09 | -1,18 |
| 43 | -1,65 | -1,49 | -1,24 | -1,31 |
| 44 | -1,43 | -1,44 | -1,75 | -1,46 |
| 45 | -2,22 | -2,19 | -1,65 | -1,74 |
| 46 | -2,14 | -2,47 | -2,99 | -2,82 |
| $r$ | 0,975 | 0,975 | 0,987 | – |
| $z$ | 2,177 | 2,177 | 2,521 | – |

Анализ данных табл. 1 показывает, что значения сил испытуемых по разным моделям не совпадают, нет между ними также и линейной зависимости. Интерес представляет установление корреляции между рядами данных. В предпоследней строке табл. 1 приведены значения коэффициента корреляции Пирсона между значениями, рассчитанными по модели, представленной в соответствующем столбце, и по предлагаемой трехпараметрической модели (принятой за основу для сравнения, поскольку она является обобщением всех рассматриваемых моделей).

Анализ значений *r* выявляет следующее: если считать, что наиболее точные значения силы испытуемого дает предлагаемая модель, поскольку она является обобщением всех остальных рассматриваемых, модель Бирнбаума с определением дифференцирующей способности заданий и модель Раша дают одинаковые по степени точности значения (*r* составляет 0,975 для обоих моделей). Модель Бирнбаума с определением дифференцирующей способности испытуемых дает наиболее близкие к предлагаемой модели значения ($r = 0,987$).

Поскольку выборочные распределения коэффициентов корреляции считаются слишком сложными для практического использования, для определения значимости сделанных на основе значений *r* выводов осуществлен переход от *r* к введенной Фишером мере *z*, имеющей, в отличие от *r*, нормальное распределение с математическим ожиданием $z = 0{,}5\ln\dfrac{1+r}{1-r}$ и дисперсией $D_z = \dfrac{1}{n-3}$, где *n* – количество испытуемых в выборке [4, т. 1, с. 61–62]. Вычисленные значения *z* приведены в последней строке табл. 1. Стандартное отклонение значений *z* составляет $\sigma_z = \sqrt{D_z} = \sqrt{\dfrac{1}{n-3}} = \sqrt{\dfrac{1}{46-3}} = 0{,}152$. Сравнение значений *z* со значением $\sigma_z$ позволяет сделать вывод, что разница между значениями *r* для моделей Раша и 1-го варианта модели Бирнбаума несущественна, а между значениями *r* для модели Раша и 2-го варианта модели Бирнбаума значима, впрочем лишь на 10%-м уровне значимости.

Так, разница между значениями *z* для модели Раша и 1-го варианта модели Бирнбаума составляет $\delta = 0$, что при значении $\sigma_\delta = \sqrt{2}\sigma_z = 0{,}216$ дает отношение $\dfrac{\delta}{\sigma_\delta} = 0$. Для случая модели Раша и 2-го варианта модели Бирнбаума $\delta = 2{,}521 - 1{,}177 = 0{,}344$, и $\dfrac{\delta}{\sigma_\delta} = \dfrac{0{,}344}{0{,}216} = 1{,}60$. 10%-граница для $\left(\dfrac{\delta}{\sigma_\delta}\right)$ примерно равна этому значению (1,64), и сделанный вывод на 10% уровне значимости подтверждается [2, с. 382].

В таблице 2 приведены рассчитанные значения силы заданий и их рангов согласно модели Раша, модели Бирнбаума и предложенной трехпараметрической модели. Задания расположены по возрастанию их силы согласно предложенной модели.

*Таблица 2*

**Значения сил заданий в рамках рассматриваемых моделей**

| № задания | Модель Раша | Модель Бирнбаума | | Предлагаемая модель |
|---|---|---|---|---|
| | | Вариант 1 | Вариант 2 | |
| **1** | **2** | **3** | **4** | **5** |
| 1 | -1,21 | -1,71 | -0,95 | -1,96 |
| 2 | -0,29 | -1,06 | -0,20 | -1,80 |
| 3 | -2,00 | -1,73 | -2,11 | -1,54 |
| 4 | -1,02 | -1,46 | -1,11 | -1,48 |
| 5 | -0,69 | -1,18 | -0,64 | -1,27 |
| 6 | -1,39 | -1,14 | -1,25 | -1,08 |
| 7 | -0,79 | -1,16 | -0,71 | -1,07 |
| 8 | -0,80 | -1,01 | -0,77 | -1,05 |
| 9 | -1,55 | -0,89 | -1,21 | -0,97 |

*Окончание табл. 2*

| 1 | 2 | 3 | 4 | 5 |
|---:|---:|---:|---:|---:|
| 10 | -0,56 | -1,08 | -0,59 | -0,96 |
| 11 | -0,98 | -0,84 | -0,77 | -0,93 |
| 12 | -1,08 | -0,67 | -0,93 | -0,69 |
| 13 | -1,18 | -0,95 | -0,94 | -0,65 |
| 14 | -0,91 | -0,46 | -0,80 | -0,50 |
| 15 | -0,57 | -0,50 | -0,61 | -0,43 |
| 16 | -0,66 | -0,66 | -0,59 | -0,38 |
| 17 | -0,22 | -0,26 | -0,11 | -0,22 |
| 18 | -0,36 | -0,26 | -0,31 | -0,22 |
| 19 | -0,22 | -0,18 | -0,38 | -0,17 |
| 20 | -0,06 | -0,03 | -0,24 | -0,03 |
| 21 | -0,09 | -0,05 | -0,16 | 0,00 |
| 22 | -0,02 | 0,00 | 0,02 | 0,08 |
| 23 | -0,02 | 0,09 | -0,11 | 0,14 |
| 24 | 0,12 | 0,14 | 0,07 | 0,14 |
| 25 | 0,05 | 0,10 | 0,00 | 0,15 |
| 26 | 0,06 | 0,15 | -0,02 | 0,15 |
| 27 | 0,20 | 0,22 | 0,00 | 0,22 |
| 28 | 0,49 | 0,31 | 0,29 | 0,29 |
| 29 | 0,42 | 0,22 | 0,32 | 0,34 |
| 30 | 0,12 | 0,36 | 0,03 | 0,36 |
| 31 | 0,31 | 0,46 | 0,17 | 0,45 |
| 32 | 0,73 | 0,48 | 0,51 | 0,47 |
| 33 | 0,57 | 0,37 | 0,49 | 0,48 |
| 34 | 0,82 | 0,60 | 0,74 | 0,66 |
| 35 | 0,23 | 0,61 | 0,22 | 0,70 |
| 36 | 0,90 | 0,82 | 0,72 | 0,77 |
| 37 | 0,40 | 0,79 | 0,24 | 0,86 |
| 38 | 1,49 | 1,14 | 1,35 | 1,07 |
| 39 | 0,53 | 1,08 | 0,31 | 1,22 |
| 40 | 1,21 | 1,32 | 1,09 | 1,35 |
| 41 | 2,07 | 1,89 | 2,20 | 1,65 |
| 42 | 1,06 | 1,59 | 0,95 | 1,80 |
| 43 | 2,48 | 2,12 | 2,76 | 1,86 |
| 44 | 2,45 | 2,42 | 3,02 | 2,18 |
| $r$ | 0,911 | 0,986 | 0,886 | – |
| $z$ | 1,532 | 2,485 | 1,402 | – |

Анализ данных табл. 2 позволяет сделать выводы, аналогичные сделанным согласно данным табл. 1, с закономерной взаимной заменой вариантов 1 и 2 модели Бирнбаума. Значения сил заданий, полученные согласно модели Бирнбаума (2-й вариант) и модели Раша, примерно равноценны. Наибольшее приближение сил заданий к рассчитанным по предлагаемой трехпараметрической модели дает 1-й вариант модели Бирнбаума. Подсчет значений $z$ и $\sigma_z = 0{,}156$ подтверждает высказанное.

Все сказанное позволяет сделать следующие выводы:

Возможно создание трехпараметрической модели тестирования с использованием в качестве параметров: 1) разности сил испытуемого и задания; 2) дифференцирующей способности испытуемого; 3) дифференцирующей способности задания.

Модель Бирнбаума дает наиболее близкие в сравнении с предлагаемой моделью значения сил той стороны тестирования (испытуемых либо заданий), чья дифференцирующая способность входит в модель. Для стороны тестирования, чья дифференцирующая способность в модели не учитывается, улучшение точности определения силы не наблюдается.

Применение предлагаемой трехпараметрической модели с одновременным определением дифференцирующей способности и испытуемых, и заданий позволяет определить значения сил испытуемых и заданий с высокой точностью.

Задействованные размеры выборок (46 по испытуемым и 44 по заданиям) позволяют сделать осторожное заключение о значимости сделанных выводов. Вместе с тем окончательное суждение о применимости предложенной модели и ее свойствах в сравнении с традиционно используемыми моделями требует проведения многократных исследований на бо́льших выборках.

## Список литературы